\documentclass[letterpaper, 10 pt, conference]{ieeeconf}  

\IEEEoverridecommandlockouts                              
\usepackage{graphicx}
\usepackage{epsfig} 
\usepackage{mathptmx} 
\usepackage{times} 
\usepackage{amsmath} 
\usepackage{amssymb}  
\usepackage{bm}  
\usepackage{comment}  
\usepackage{amsbsy}
\usepackage{lipsum}
\usepackage{multirow}
\usepackage{booktabs}
\usepackage{url}

\usepackage{kotex}
\usepackage{url}
\usepackage[ruled,vlined,linesnumbered]{algorithm2e}
\usepackage[dvipsnames]{xcolor}

\title{\LARGE \bf
Online Trajectory Generation of a MAV for Chasing a Moving Target in 3D Dense Environments}

\author{Boseong Felipe Jeon and H. Jin Kim
\thanks{*This material is based upon work supported by the Ministry of Trade, Industry \& Energy(MOTIE, Korea) under Industrial Technology Innovation Program. No.10067206, 'Development of Disaster Response Robot System for Lifesaving and Supporting Fire Fighters at Complex Disaster Environment'}
\thanks{Department of mechanical and aerospace engineering,
        Seoul national university of South Korea
        {\tt\small \{a4tiv,hjinkim\}@snu.ac.kr}}%
}

\begin{document}

\maketitle
\thispagestyle{empty}
\pagestyle{empty}

\begin{abstract}
This work deals with a moving target chasing mission of an aerial vehicle equipped with a vision sensor in a cluttered environment. In contrast to obstacle-free or sparse environments, the chaser should be able to handle collision and occlusion simultaneously with flight efficiency. In order to tackle these challenges with real-time replanning, we introduce a metric for target visibility and propose a cascaded chasing planner. By means of the graph-search methods, we first generate a sequence of chasing corridors and waypoints which ensure safety and optimize visibility. In the following phase, the corridors and waypoints are utilized as constraints and objective in quadratic programming from which we complete a dynamically feasible trajectory for chasing. The proposed algorithm is tested in multiple dense environments. The simulator \textit{AutoChaser} with full code implementation and GUI can be found in \url{https://github.com/icsl-Jeon/traj_gen_vis}.
\end{abstract}

\section{INTRODUCTION}

Due to the enhanced autonomy from localization \cite{LOC1,LOC2}, tracking \cite{TRACK} and control \cite{CTRL}, MAVs have been widely utilized for the vision-based tasks such as videography and surveillance. Especially, planners proposed in the previous research (\cite{FOV,INTERCEPT,OH}) improved the performance of autonomous target following  with real-time applicability.
Although those methods can be reasonably employed in obstacle-free or sparse situations, there remain issues for the general cluttered environments.

The presence of multiple obstacles poses two main difficulties in the chasing task: collision and occlusion which should be handled simultaneously. Aside from the safety consideration, occlusion handling especially becomes critical in a situation where a chaser does not have perfect information on the future movement of a target. In the case, the target visibility over the future time horizon is not guaranteed although it is visible from chaser at current time. Additionally, if the chaser spends too much resource (e.g. travel distance) in preparation for the visibility in the future horizon, inefficient motion will be caused. 

To overcome the issues, a chasing planner needs to optimize visibility and travel efficiency with safety guarantees. Also, fast computation is essential to effectively respond to the uncertainty of target movement.              
\begin{figure}[t]
\label{fig:intro}
\centering
\includegraphics[width=0.45\textwidth]{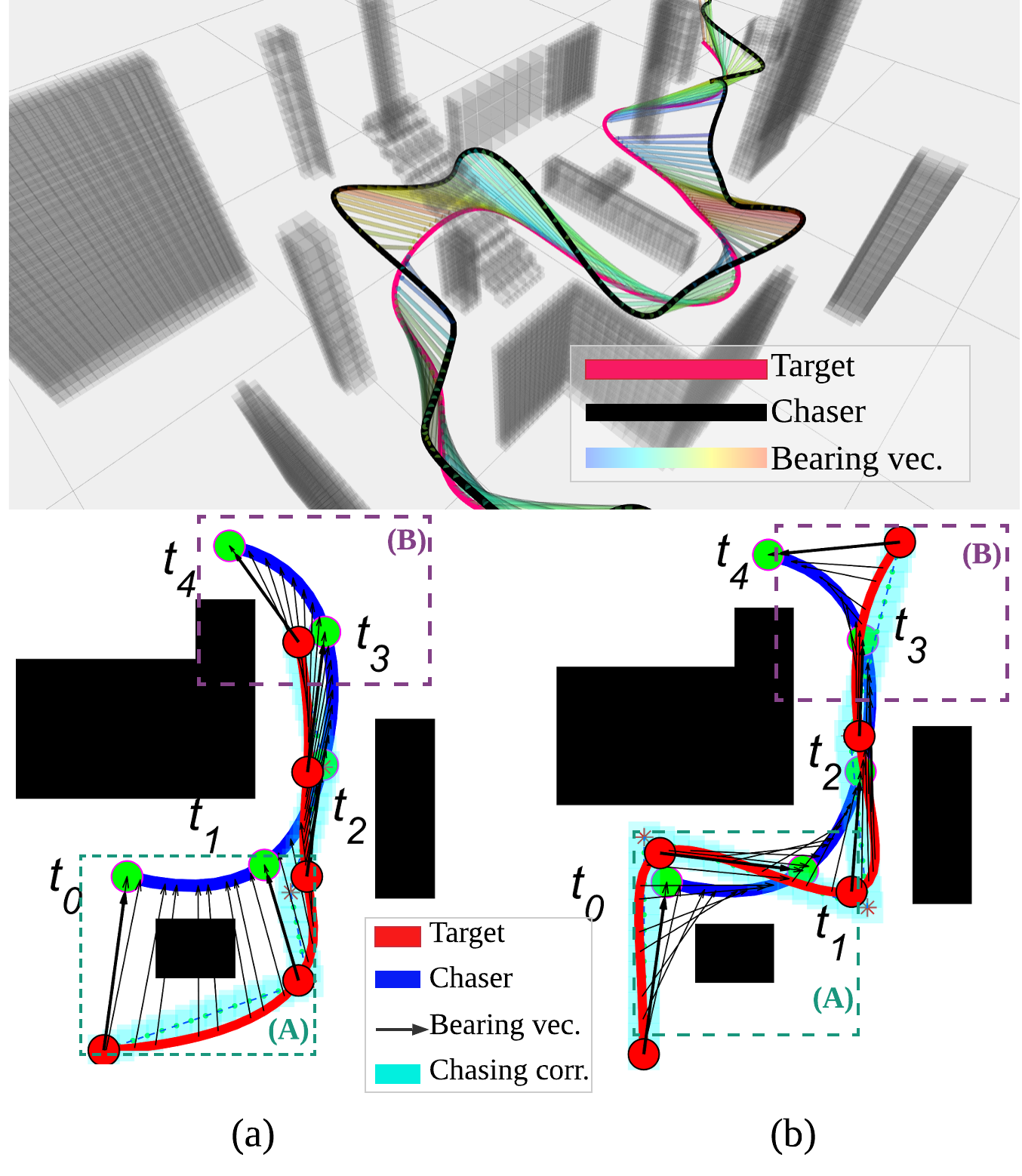}

\caption{Upper: chasing mission in \textit{complex city} example. The target moves on the ground following red-coloured line. Bearing vectors connecting the target and chaser are stamped with the visibility score in the jet colormap scale (Sec. \ref{metric},\ref{sim}). Lower: chasing trajectories obtained from different weights of visibility $w_v$ (Sec. \ref{preplanning}). In (b), $w_v$ is set to a larger value than (a). The chasing corridor (cyan) can be adjusted based on the relative importance of travel efficiency and visibility while ensuring safety. As depicted in dashed boxes (A),(B), different chasing behaviors were obtained for different $w_v$. In case of lower $w_v$, the target was not visible in dashed box region (A) although the chaser travels with shorter distance. The strategy of higher $w_v$ efforts to achieve a more safe viewpoint against the uncertain target movements after $t_4$ (box region (B)).}
\end{figure}
 In this paper, we propose an online chasing planner which handles occlusion and collision in an integrated way against dense obstacles. 
 
 Our key contributions are twofold. Firstly, we introduce a visibility metric for a moving target which can be defined in general environments without an restriction on the shape of obstacles. With a simple computation, the metric encodes the sensitivity of visibility against uncertainty of future movement of the target. Secondly, a cascaded chasing strategy with real-time performance is presented. Rather than adopting non-convex optimization with multiple objectives, we seed sparse viewpoints which guarantee safety and visibility by means of an graph-search method. Based on the discrete viewpoints skeleton, piecewise polynomials are generated in quadratic programming (QP) formulation with additional safe consideration. The planner guarantees the safety of the entire chasing path while adjusting the importance between travel efficiency and visibility. We found that the cascaded planner ensures the chasing quality of trajectory without sacrificing the online computational speed.

\section{RELATED WORKS}

\subsection{Chasing in obstacle environments}
The research  \cite{ARM}, \cite{SWITCH} dealt with the occlusion and collision in chasing task based on a formulation in image space.  \cite{ARM} took advantage of quadratic programming (QP) to compute a control input to avoid occlusion from a single polyhedral occluder. A switching controller was used in  \cite{SWITCH}, which is composed of multiple objectives (a nominal tracking, collision avoidance and occlusion handling). Although the formulations in the mentioned works do not require the reconstruction of target in 3D space, the methods were validated either in a single obstacle or sparse environments.

\label{PBVS}
In contrast to \cite{ARM}, \cite{SWITCH}, a group of research \cite{CINE1,CINE2,HKUST,SENSOR} used Cartesian space formulation. Those works employed 3d reconstruction process to deal with more complex scenarios by the virtue of reliable mapping and localization. To account the occlusion into consideration, the authors in \cite{SENSOR} introduced a risk function which is defined as the shortest distance for the target to escape the sensor range of chaser to evaluate the visibility. Although the algorithm was validated against multiple obstacles, it has to calculate the precise geometry of obstacles for occlusion culling. \cite{HKUST} employed safe corridor generation method \cite{SF3} for collision avoidance for dynamic target chasing. They examined the trade-off between flight efficiency and collision avoidance. Despite of safe flight avoiding obstacles, visibility for a target was not considered, without guarantee for maintaining target in the sight. 

More recently, \cite{CINE1,CINE2} proposed planning strategies which consider occlusion and collision simultaneously in obstacle environments. \cite{CINE1} defined the cost functions for visibility and safety consideration based on the assumption of the ellipsoidal shape of occluder. \cite{CINE2} formulated an optimization problem under octomap framework (\cite{OCTOMAP}) accommodating more general shape of obstacles. In order to measure visibility of a target, the authors calculated the integral of signed distance field (SDF) over a 2D manifold. 


The works \cite{CINE1,CINE2} both defined nonlinear cost function and optimized it with gradient descent methods.
Although these approaches guarantee the dynamic feasibility of MAVs and compute the solution online, they are subject to the local optimal due to non-convexity of multi-objectives cost function to account for safety and visibility simultaneously. It is especially ineffective to entirely rely on the non-convex formulation in more dense environments as pointed out in \cite{RRT-PP}.  

\subsection{Safe trajectory generation with pre-planning}
In this subsection, we mention several recent research on safe planning to reach a static goal while avoiding obstacles. Although its mission is different from ours, it is worthy to note the approaches developed in  \cite{SF2,SF3,RRT-PP} to deal with dense environments. The works adopted a cascaded planner where a sequence of sparse waypoints or together with safe corridor is seeded in the first phase, which is followed by continuous trajectory completion. Let us call the first process as \textit{preplanning}. The discrete optimization such as search-based \cite{JPS} or sampling based methods \cite{RRT} were used for the preplanning phase. The output of it acts as a good initial guess or constraints for the continuous optimization circumventing the local solutions of poor quality. 
The applicability of the cascaded planners in the mentioned works were validated in complex environments enjoying safety, dynamic feasibility and stability in optimization. Inspired by the approaches, we develop a preplanner for the autonomous chasing which is effective for the more challenging cases.

\begin{figure}[t]
\centering
\includegraphics[width=0.4\textwidth]{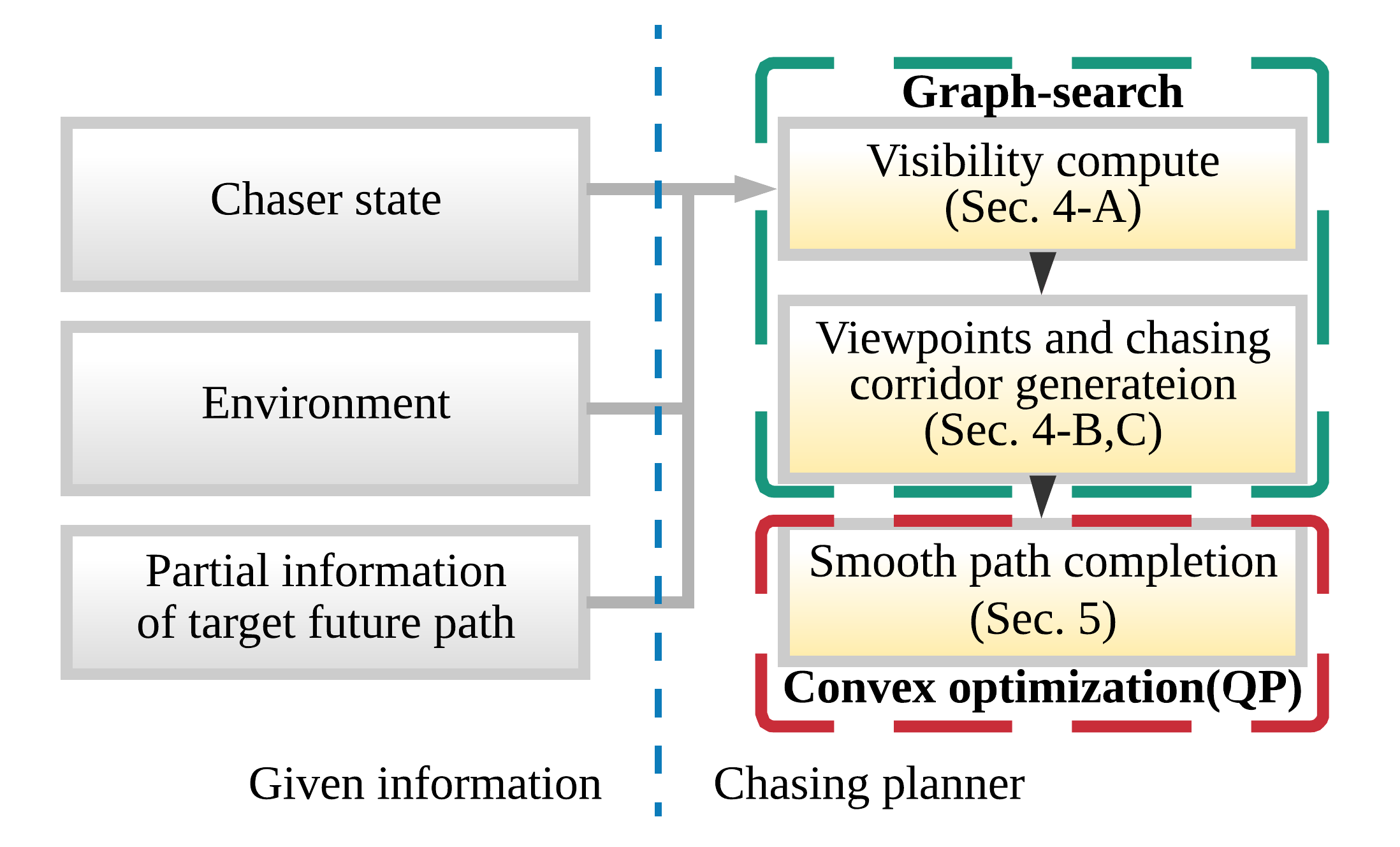}
\caption{Overview of the planning pipeline for the proposed chasing strategy. This work focuses on the motion planner which consists of two phases (i.e. two boxes in the right side) in response to the update of information. The preplanning process (green dashed box) generates the sequence of viewpoints which are safe and visible. The lines connecting each point also guarantee the safety clearance from obstacles, from which a set of corridors are obtained. Based on the corridors and viewpoints, a smooth chasing path is computed with convex optimization (QP).}     
\label{fig:overview}
\end{figure}
  
\section{PROBLEM STATEMENT}

\subsection{Problem settings}
\label{assumptions}
In our chasing setting, multiple obstacles of arbitrary shape are involved. The environment is assumed to be known as a priori. MAV (chaser) is assumed to be equipped with a vision sensor with finite FOV for the purpose of target tracking and state estimation. In this paper, these processes are assumed to be reliable (\cite{LOC1,TRACK}) and we focus on the motion strategy for chasing. Regarding a moving target of interest, it is assumed to have bounded accelearation, and is located in the FOV of the MAV initially. The entire path of the target is not available a priori to the chaser. Instead, the chaser is provided with the future movement of the target over only short horizon, which is repeatedly updated in the air in a similar manner with \cite{INTERCEPT,HKUST}. 
         
\subsection{Capabilities}
\label{subsec:cap}
Based on these assumptions on the chaser and target, we aim to build a chasing planner which has the following capabilities. 
\begin{enumerate}
            \item Safety of entire chasing path is guaranteed.
            \item Visibility of target is guaranteed at pre-defined time steps.
            \item Visibility of target and total travel distance is optimized. 
            \item The integral of high order derivatives of chasing path is minimized for the smoothness and flight efficiency of the path.
            \item Real-time performance of planner is achieved in response to the online information updated  (i.e. the left side in Fig. \ref{fig:overview}).   
\end{enumerate}

\subsection{Nomenclature}

\newcommand{\configspace}{\mathbf{\chi}}
\newcommand{\free}{\mathbf{\mathbf{\chi}}_{free}}
\newcommand{\obs}{\mathbf{\mathbf{\chi}}_{obs}}
\newcommand{\vis}{\mathbf{\mathbf{\chi}}_{vis}}
\newcommand{\occ}{\mathbf{\mathbf{\chi}}_{occ}}
\newcommand{\dobs}{\mathbf{\mathbf{\chi}}_{dobs}}
\newcommand{\docc}{\mathbf{\mathbf{\chi}}_{docc}}
\newcommand{\ax}{{{x}}}
\newcommand{\state}{{{s}}}
\newcommand{\R}{{\mathbb{R}}^{3}}
\newcommand{\RS}{{\mathbb{R}}^{6}}
\newcommand{\visfield}{\psi(\ax;\ax_p)}
\newcommand{\EDT}{\mathrm{EDT}(\ax;\obs)}
\newcommand{\seq}{{\sigma}_{[t_0,t_N]}}
\newcommand{\transiscore}{c_v(\ax_{n-1},\ax_{n})}
\newcommand{\visscore}{\psi_v(\ax;\state_p)}

Before preceding, we define several notations for the ease of discussion.
\begin{itemize}
    \item[--] $\ax_{c}\in\R$ : Position of a chaser.
    \item[--] $\ax_{p}\in\R$ : Position of a target.
    \item[--] $L(\ax_{1},\ax_2) = \{\ax|\; t\ax_1+(1-t)\ax_2,\  0 \leq t \leq 1\}$ : The line segment connecting $\ax_1,\ax_2 \in \R$. 
    \item[--] $B(p,{l}) = \{\ax|-{l}\leq\ax-p\leq{l}\} \ (p, {l}\in \R)$ : The set of points in the axis-parallel 3D box represented by scale ${l}$ and center position $p$.
    \item[--] $\mathbf{\chi}\subset\R $ : Configuration space.
    \item[--] $\mathbf{\chi}_{free} = \{{\ax}|\;P({\ax})<\epsilon\}$ : Free space in $\configspace$, i.e. the set of points where the probability of occupancy $P({\ax})$ obtained from octomap is very small. 
    \item[--] $\mathbf{\chi}_{obs} = \configspace \setminus \free$ : Space occupied by obstacles. 

    \item[--] $\mathbf{\chi}_{vis}(\ax_{p}) = \{\ax|\; L(\ax,\ax_{p}) \cap \obs = \emptyset \} $ : A set of visible vantage points for a target position $\ax_p$.  
    \item[--] $\mathbf{\chi}_{occ}(\ax_{p}) = \configspace \setminus \vis $ : A set of occluded points for a target position $\ax_p$.

\end{itemize}

\section{VIEWPOINTS GENERATION}
\subsection{Visibility metric}
\label{metric}
In this subsection, we explain the metric for visibility of $x_p$ to be maximized for the robustness of chasing. 
 We assume that the chaser has map information $\configspace$ as stated in the Sec. \ref{assumptions}.   For simplicity,  we assume that the chaser takes $\ax_p - \ax_c$ as its bearing vector.
 
 We first consider the level set function $\phi(\ax)$ for $\ax \in \R$ to represent the obstacle region $\obs$ subject to $\phi(\ax)\leq 0$ for $\ax \in \obs$ and $\phi(\ax)>0$ for $\ax \in \free$. $\phi(\ax)$ is chosen as Euclidean distance field \texttt{EDF}($\ax;\obs$) which can be computed efficiently by the methods such as \cite{EDT1}. Based on $\phi(\ax)$, we define the visibility of target  $x_p$ when observed from the position $x$ with bearing vector $x_p - x$ as below.
\begin{equation}
\label{visfield}
    \visfield = \underset{L(\ax,\ax_p)}{\text{min}} \phi(\ax)
\end{equation}

\begin{figure}[t]
\centering
\includegraphics[width=0.5\textwidth]{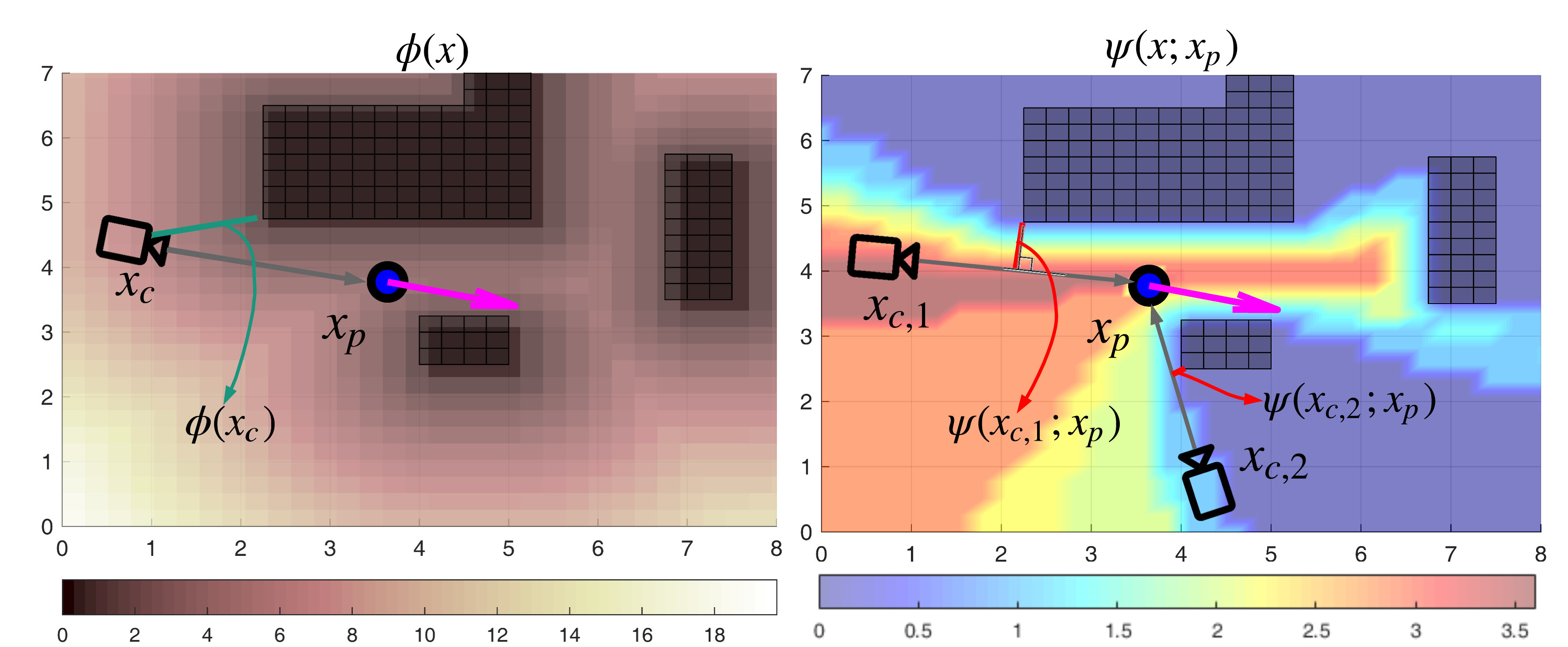}

\caption{Grid field of $\phi(x)$ (left) and $\psi(\ax;\ax_p)$ (right) in a 2D case. In the figure, a target position $x_p$ is denoted as a blue circle with velocity(magenta quiver) and chaser $x_c$ as a camera icon. $\phi(x)$ corresponds to the closest distance to the ambient obstacles while $\psi(x;x_p)$ is the closest distance between bearing vector $x_p - x_c$ and obstacles as the red lines show in the right figure. Following the definition, chaser $x_{c,1}$ secures higher visibility score than $x_{c,2}$ indicating that $x_{c,1}$ is more robust for observation against the future movement of $x_p$. We will use $\phi(x)$ as a measure for collision safety and $\psi(x;x_p)$ as visibility of $x_p$.}
\label{fig:metric_dynamic}
\end{figure}

The interpretation of (\ref{visfield}) is the closet distance between the line segment $L(\ax,\ax_p)$ and elements in $\obs$ because $\phi(\ax)$ is equivalent to \texttt{EDF}($\ax;\obs$) as illustrated in Fig \ref{fig:metric_dynamic}. Because (\ref{visfield}) is a level set representation for the occlusion region and visible region with respect to $x_p$, $\vis(\ax_p)$ can be expressed as a set $\{\ax|\psi(\ax;\ax_p)>0\}$ and $\occ(x_p)$ with  $\{\ax|\psi(\ax;\ax_p)\leq0\}$. 
The concept of the visibility level set (\ref{visfield}) was used in several literature for occlusion rendering \cite{levelset4} to compute the visible volume of environment. As we focus on a chasing mission, we will leverage $\visfield$ to gauge how reliably a chaser position $\ax$ can maintain visibility for the target $\ax_p$. 
\begin{figure*}[t]
\label{fig:preplanning}
 \center

  \includegraphics[width=\textwidth]{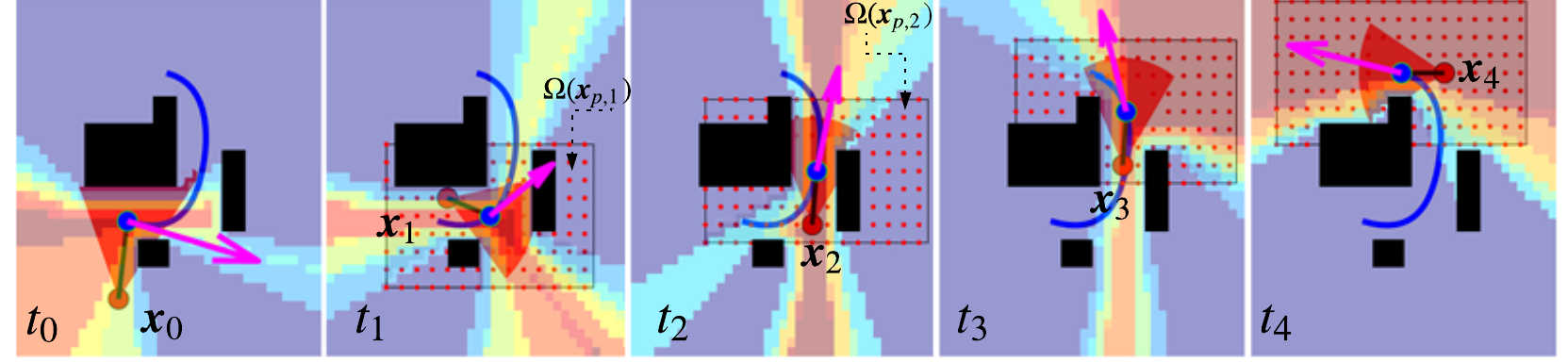}
  
  \caption{An illustrative preplanning process for chasing waypoints ${\sigma}_{[t_0,t_4]}$ in a 2D case: The thick red circle denotes $x_c$ and blue circle means $x_p$. The set $\Omega({\state}_{p,n})$ are depicted as a set of red points. To obtain ${\sigma}_{[t_0,t_4]}$, The waypoint $x_{n}$ for $t_n$ is chosen as one of the elements of $\Omega({\state}_{p,n})$ by means of graph search. For the graph construction, we have nodes $v_n\in\Omega({\state}_{p,n})$ connected with edges $e=(v_n,v_n+1)$ based following \textbf{Alogrithm 1}.}

\end{figure*}

In this paper, we operate on a 3D grid map $D$ for the compatibility with octomap. The definition of visibility (\ref{visfield}) is suitable for our scenario for the following reasons: 
First,  (\ref{visfield}) can be computed from $D$ that represents  a more general $\configspace$ than the previous research on chasing where particular shapes of obstacles are assumed (e.g. ellipsoids in  \cite{CINE1} and polygons in \cite{SENSOR}).  Secondly, the computation of (\ref{visfield}) is tractable for online applications. In our problem setting, (\ref{visfield}) can be calculated within order of milliseconds once $ \phi(\ax)$ is obtained. Thirdly, (\ref{visfield}) offers a measure of  the  robustness of a viewpoint  for $x_p$, because $\psi(x;x_p)$ is level set function encoding the sensitivity for occlusion with respect to the perturbation of $\ax$ or $\ax_p$ as discussed in \cite{levelset3} (please refer Fig. \ref{fig:metric_dynamic}). This helps us reliably evaluate the quality of viewpoint against unexpected future movements of target.



\subsection{Preplanning with safety and visibility}
\label{preplanning}

Utilizing $\phi(x)$ and $\psi(x;x_p)$ as measures for safety and visibility of target respectively, a \textit{preplanner} is introduced to seed a sequence of waypoints in response to the available information of the target's future path for a short horizon $H$. As mentioned in Sec. \ref{assumptions}, the chaser at the current position $\ax_{c}(t)$ is provided with $x_p(\tau)$ $(t\leq\tau\leq t+H)$. Let
us divide the time horizon with time step $\Delta t = \frac{H}{N}$, and let 
${\sigma}_{[t_0,t_N]}=(\ax_0,\ax_1,...,\ax_N)$ denote a sequence of $N+1$ waypoints at where $\ax_n = \ax_c(t_n)$ and $t_n=t+n\Delta t$ ($n = 0,1,..,N$). 

 For $x_{p,n}=x_p(t_n)$, we now define the \textit{transitional visibility cost} $\transiscore$ between the two wapoints $\ax_{n-1}$ and $\ax_{n}$ as below. 
\begin{equation}
    \label{eq:transitvis}
\begin{aligned}
     & \transiscore = \\& \Bigg(\sqrt{\int_{L(\ax_{n-1},\ax_n)} \psi(\ax;x_{p,n-1})d\ax \int_{L(\ax_{n-1},\ax_n)} \psi(\ax;x_{p,n})d\ax}\Bigg)^{-1}       
\end{aligned}
\end{equation}
Eqn. (\ref{eq:transitvis}) represents inverse of geometric mean of the line integrals of the two successive visibility fields corresponding to $x_{p,n-1}$ and  $x_{p,n}$ and $\transiscore$ becomes $+\infty$ if all the points along $L(\ax_{n-1},\ax_{n})$ belong to $\occ(x_{p,n})$ or $\occ(x_{p,n-1})$.  
Our goal is to find  a preplanning path ${\sigma}_{[t_0,t_N]}$ which is the solution of the following optimization problem.     
\begin{equation}
\label{eqn: optim for preplanning}
\begin{aligned}
& \underset{{\sigma}_{[t_0,t_N]}}{\text{argmin}}
&& \sum_{n=1}^{N}c(\ax_{n-1},\ax_n) \\& \mathrm{where} & &c(\ax_{n-1},\ax_n)=\underbrace{ \lVert {x}_{n-1} -{x}_{n}\rVert^2}_{\mathrm{interval \:distance}} \: + \underbrace{w_{v} \; {c_{v}}({x}_{n-1},{x}_{n})}_{\mathrm{visibility}}  \\ & & & + w_d \underbrace{(\lVert {{x}}_{p,n} -{x}_{n}\rVert - d_{des})^{2}}_{\mathrm{tracking\: distance}} \\
& \text{subject to} & &\text{${x}_{0} = \ax_{c}(t_0)$} \\
& & & \ax_n \in \vis({\ax}_{p,n}) \cap \Omega(x_{p,n})  \\
& & &\underset{\ax \in L(\ax_{n-1},\ax_{n})}{\text{min}}\phi(\ax) \geq r_{safe} \\
& & & \text{$\lVert {x}_{h-1} -{x}_{h}\rVert \leq d_{max}$} 
\end{aligned}
\end{equation}

Here, 
$r_{safe}$ is the safety clearance from obstacles and $d_{des}$ is the desired relative distance between the target and tracker. $d_{max}$ is the maximally allowable inter-distance between waypoints. For the tracking distance limits $d_{upper}, d_{lower}$ subject to $d_{lower} \le \lVert x_n-x_p \rVert  \le d_{upper}$, $\Omega(x_{p,n})$ is a set of discretized points selected in a box region $B(x_{p,n}, 2d_{upper})\setminus B(x_{p,n}, 2d_{lower})$ where a viewpoint at the corresponding time step $t_n$ will be chosen (see the set of red dots in Fig. \ref{fig:preplanning}). By adjusting the weights $w_v$ and $w_d$, a different behavior of chaser is achieved while satisfying the constraints in (\ref{eqn: optim for preplanning}). In this work, we are interested in examining the trade-off between the total travel distance and visibility, so will analyze the effect of $w_v$ in more detail (see the lower one in Fig. \ref{fig:intro} and Fig. \ref{fig:total result}). 



In order to solve  (\ref{eqn: optim for preplanning}), we now construct a directed graph $G(V,E)$ where each node corresponds to a candidate waypoint for ${\sigma}_{[t_0,t_N]}$. For a vertex set 
$V =\bigcup\limits_{n=0}^{N+1} V_n$ where  
\begin{equation}
V_n = 
\begin{cases}
\{\ax_c(t_0)\} & n=0 \\
\Omega(x_{p,n}) & 1 \leq n \leq N \\
\{x_{dummy}\} & n = N+1
\end{cases}
\end{equation}
$x_{dummy}$ is introduced to wrap the graph as a virtual goal point as the specific goal point is not pre-defined in our case. We construct $E$ by following process in \textbf{Algorithm 1}(the notation in \textbf{Algorithm 1} is based on the operators used in C++).
The solution of (\ref{eqn: optim for preplanning}) is computed by means of the shortest path search methods on $G$ from $\ax_c(t_0)\in V_0$ to $x_{dummy}\in V_{N+1}$.

\subsection{Chasing corridor generation}
\label{comp_discuss}
 Based on the solution ${\sigma}_{[t_0,t_N]}$ obtained from the previous section, we generate a set of chasing corridors to be used for computing a continuous optimal trajectory which is described in the next section. For the moment, let us consider the following linear interpolation  ${\gamma}(\tau;{\sigma}_{[t_0,t_N]})$ for $t_0\leq\tau\leq t_N$: 
\begin{equation}
    {\gamma}(\tau;{\sigma}_{[t_0,t_N]}) = 
    \dfrac{\tau-t_{n}}{\Delta t}\ax_{n-1}+\dfrac{\tau-t_{n-1}}{\Delta t}\ax_{n}\:(t_{n-1}\leq\tau\leq t_{n})
\end{equation}

Owing to our formulation (\ref{eqn: optim for preplanning}), ${\sigma}_{[t_0,t_N]}$ satisfies ${\gamma}(\tau;{\sigma}_{[t_0,t_N]})\in \free $ for $t_0\leq\tau\leq t_N$ and ${\gamma}(t_n;{\sigma}_{[t_0,t_N]})\in \vis({\ax}_{p,n})$ for $n = \{0,1,..,N\}$ as safety and visibility constraint as stated in the \ref{subsec:cap}-1) and 2). 
 Also, in the case of large enough weight $w_v$ and small enough $\Delta t$, we observed that the condition ${\gamma}(\tau;{\sigma}_{[t_0,t_N]})\in \vis({\ax}_p(t))$ for $t_0\leq\tau\leq t_N$ is satisfied instead of longer travel distance (see the right in Fig \ref{fig:intro}).   
 
 We now define the chasing corridor at $\tau$ as below:
\begin{equation}
\label{corridor}
    Q(\tau;{\sigma}_{[t_0,t_N]}) = B(\gamma(\tau;{\sigma}_{[t_0,t_N]}),{l}_{corr}(\tau)),  
\end{equation}
where the parameter 
 ${l}_{corr}(\tau)\in \R$ representing the size of a corridor at $\tau$ is selected to satisfy the following necessary condition for the  safety: $({l}_{corr}(\tau))_{i} \leq \dfrac{\phi(\gamma(\tau;{\sigma}_{[t_0,t_N]})}{\sqrt{3}}$ for $i = 1,2,3$. As a side note, although relaxing ${l}_{corr}$ may be advantageous for the flexibility of smooth path generation, it can reduce the visibility performance between the two points ($x_{n-1},x_{n}$). Examples of the chasing corridor are shown in the lower part of Fig. \ref{fig:intro} for the two different values of $w_v$.


\begin{algorithm}
\DontPrintSemicolon
\SetAlgoLined
\SetKwInOut{Input}{Input}\SetKwInOut{Output}{Output}
\SetKwInOut{given}{Given}
\SetKwInOut{param}{Parameter}
\SetKwInOut{initialize}{Initialize}
\SetKwFunction{find}{find}
\SetKwFunction{append}{append}
\SetKwFunction{Edge}{Edge}
\given{$\phi(\cdot)$, $c(\cdot,\cdot)$, $V_n$, $x_{p,n}$}
\BlankLine
\initialize{$V =\bigcup\limits_{n=0}^{N+1} V_n, E=\emptyset$}
\For{$n=1$ \KwTo $N+1$}{
    \ForAll{$v_{n-1} \in V_{n-1}$}{
        \ForAll{$v_{n} \in V_{n}$}{
            $\ax_{n-1}=v_{n-1} \to x$, $\ax_{n}=v_{n} \to x $ \\
            \eIf{$n<=N$}{
            \texttt{isLineSafe} $=\underset{\ax \in L(\ax_{n-1},\ax_{n})}{\text{min}}\phi(\ax) \geq r_{safe}$\\
            \texttt{isVisible1} $=\psi(\ax_{n-1};\ax_p) > 0$\\
            \texttt{isVisible2} $=\psi(\ax_{n};\ax_p) > 0$\\
            \texttt{isClose} $=\lVert\ax_{n-1} - \ax_{n}\rVert < d_{max}$ \\ 
            \texttt{isConnectable} = \texttt{isLineSafe} \& \texttt{isVisible1} \& \texttt{isVisible2} \& \texttt{isClose}  \\
            }
            {
            // connect all $v_N$ with $x_{dummy}$\\

            \texttt{isConnectable} = true
            }
            \If{\texttt{isConnectable}  }{
             
                $e  = \Edge{}$, $e\to x_1$ = $\ax_{n-1}$, $e\to x_2$ = $\ax_{n}$\\
                \eIf{$n<=N$}{
                    $e\to w = c(\ax_{n-1},\ax_{n})$\\
                }{
                // arbitrary positive number\\
                    $e\to w = positive\; const$\\  
                }
                $E\to \append{$e$}$                
            }
        }
    }
}
\Return{G(V,E)}
\caption{BuildGraph}
\end{algorithm}

\section{SMOOTH TRAJECTORY GENERATION}
\label{smoothpath}
The previous section explained how the preplanner yields a sequence of viewpoints and chasing corridors for the consideration of safety and visibility. By the virtue of the differential flatness of a quadrotor MAV \cite{MINISNAP} together with $\sigma_{[t_0,t_N]}$ and $Q(\tau;{\sigma}_{[t_0,t_N]})$ from the preplanner,  we describe how to generate a dynamically feasible path for an aerial chaser $x_c(\tau)\; (t_0\leq \tau \leq t_N)$. 

The yaw angle of the chaser is determined so that the x-axis of body frame of MAV heads to the target.  The chaser path for position is represented as the following piecewise polynomials.
\begin{equation}
x_c(\tau) =  
\begin{cases}
x_{c,1}(\tau) = \sum_{k=0}^{K}p_{1,k}\tau^k & (t_{0} \leq \tau < t_{1}) \\
x_{c,2}(\tau) = \sum_{k=0}^{K}p_{2,k}\tau^k & (t_{1} \leq \tau < t_{2}) \\
...&\\
x_{c,N}(\tau) = \sum_{k=0}^{K}p_{N,k}\tau^k & (t_{N-1} \leq \tau < t_{N}) \\
\end{cases}
\end{equation}
Here,  $p_{n,k} \in \R$ are  coefficients for the polynomial of order $K$.  They are obtained from the following optimization.

\begin{equation}
\label{eqn:cont opti}
\begin{aligned}
& \underset{p_{n,k}}{\text{min}}
&& \sum_{n=1}^{N}\Bigg( \int_{t_{n-1}}^{t_{n}} {\lVert{{x_c}^{(3)}(\tau)}\rVert}^2 d\tau \:+\: \lambda{\lVert{x_c(t_{n})}-x_{n}\rVert}^2 \Bigg)  \\
& \text{Subject to} & &  x_c(t_{0}) = x_{0}  \\
& & &   \dot{x}_c(t_{0}) = \dot{x}_{0}\\
& & &   \ddot{x}_c(t_{0}) = \ddot{x}_{0}\\
& & &   x_{c,n}^{(k)}(t_{n}) = x_{c,n+1}^{(k)}(t_{n}) \; (k= 0,1,2) \; \\ 
& & &   x_c(\tau) \in Q(\tau;{\sigma}_{[t_0,t_N]}) 
\end{aligned}
\end{equation}
Here, 
$x_0, \dot{x}_{0}, \ddot{x}_{0}$ are the initial state of the chaser when replanning is triggered at $\tau=t_0$. The fourth constraint  accounts for the continuity condition up to the second order derivatives. The last box constraint is the corridor condition at the subsampled time step $\tau = \tau_{n,i}$ between every two knots ($x_{n-1},\; x_{n}$), i.e. $t_{n-1} \leq \tau_{n,1} \leq \tau_{n,2} \leq ... \leq \tau_{n,M} \leq t_n$ where $M$ should be adjusted depending on $d_{max}$,  the prediction horizon $H$ and the polynomial order $K$. 
 
The objective of (\ref{eqn:cont opti}) is the weighted sum of integral of the jerk squared and squared error of waypoints. (\ref{eqn:cont opti}) can be transformed into quadratic programming following a similar manner to  \cite{MINISNAP,RRT-PP}. The convex optimization can be solved efficiently with algorithms such as interior-point or active-set methods. In this way, the consideration for safety, visibility and travel distance is implicitly included as a box constraint $Q(\tau;{\sigma}_{[t_0,t_N]})$ rather than including them in a non-convex objective function as  \cite{CINE1,CINE2} did.

\section{VALIDATION RESULT}

In order to implement the chasing planner in C++, BOOST graph library (BGL) was used to obtain $\sigma_{[t,t+H]}$ (as a reminder, $t$ and $t+H$ are equivalent to $t_0$ and $t_N$ respectively.) in (\ref{eqn: optim for preplanning}) by the shortest path search (Dijsktra algorithm). qpOASES \cite{QP} was employed as a QP solver for the smooth path completion in (\ref{eqn:cont opti}) and dynamicEDT3D library for computing $\phi(x)$ \cite{OCTOMAP}.  All computation was performed on a 2.7Ghz intel i7 CPU labtop.  

\begin{figure}[h]
 \centering
  \includegraphics[width=0.5\textwidth]{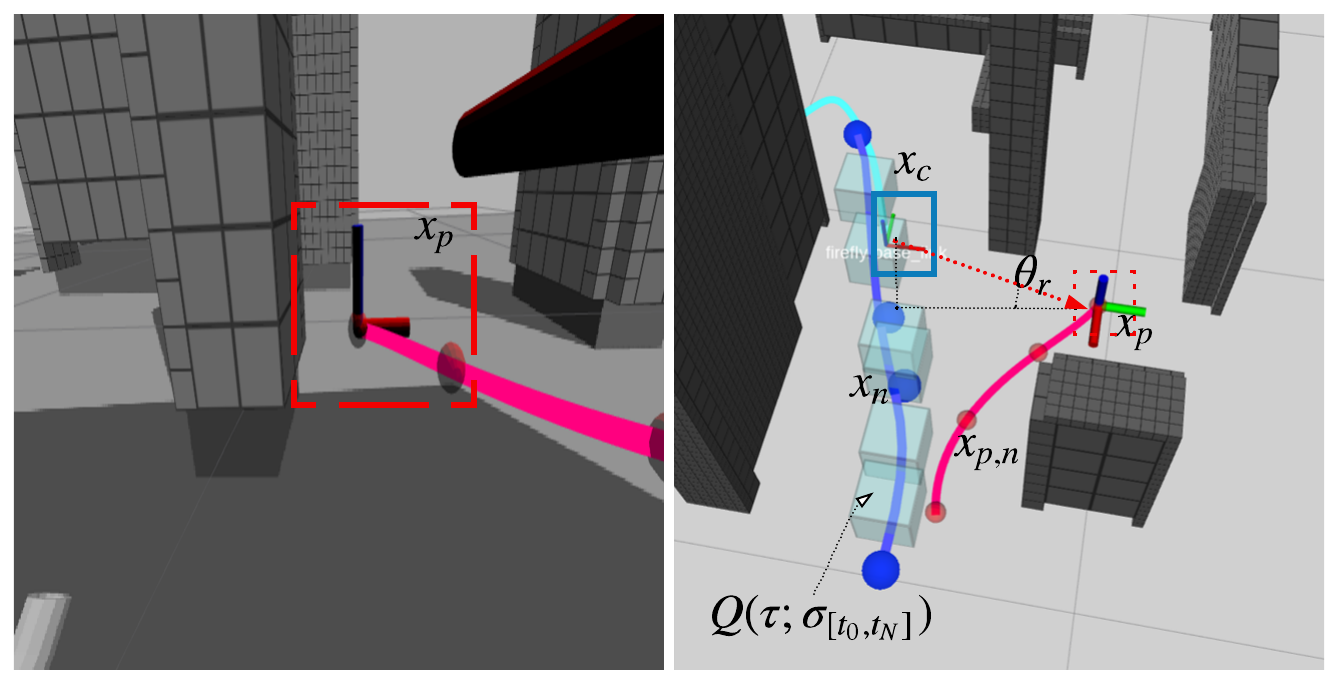}
  
  \caption{Left: image view of the target. Right: replanning result (blue line) based on the partial future information of the target $x_p(\tau)$ for $t\leq\tau\leq t+H$ (solid red-colored line). The elevation of bearing vector $\theta_r$ is depicted.}
  \label{fig:sim_image}
\end{figure}

 We tested the proposed chasing planner for two examples: \textit{mini garden} and \textit{complex city} with respect to the two different weights on the visibility. Their 3D models can be found in the first row of Fig. \ref{fig:total result}. The common parameters for the both cases are summarized in Table \ref{Table:params}.  We plotted the overall chasing result in Fig. \ref{fig:total result} while the quantitative analysis is described in Table \ref{Table:result} and Fig. \ref{fig:data}. For the realistic simulation, Rotors simulator was used \cite{ROTORS} to operate a MAV equipped with fixedly attached vi-sensor ($20^\circ$ downward and $150^\circ$ FOV). In both cases, the target is ground vehicle. 
To maintain the ground target within the view of the fixedly mounted vision-sensor, the elevation angle $\theta_r$ of the vector $x_c - x_p$ is limited between $20^{\circ}$ and $70^{\circ}$ from the consideration of mounting angle of camera and FOV. This naturally imposes the maximally allowable attitude of the chaser MAV. We assume that the chaser executes the replanned path for $t<\tau<t+H-\epsilon$ where $\epsilon>0$ after which another replanning session will be triggered. The invoked times for replanning are denoted as vertical lines in the background of Fig. 6, 7.    

\begin{figure}[h]
\centering
    \includegraphics[width=.5\textwidth]{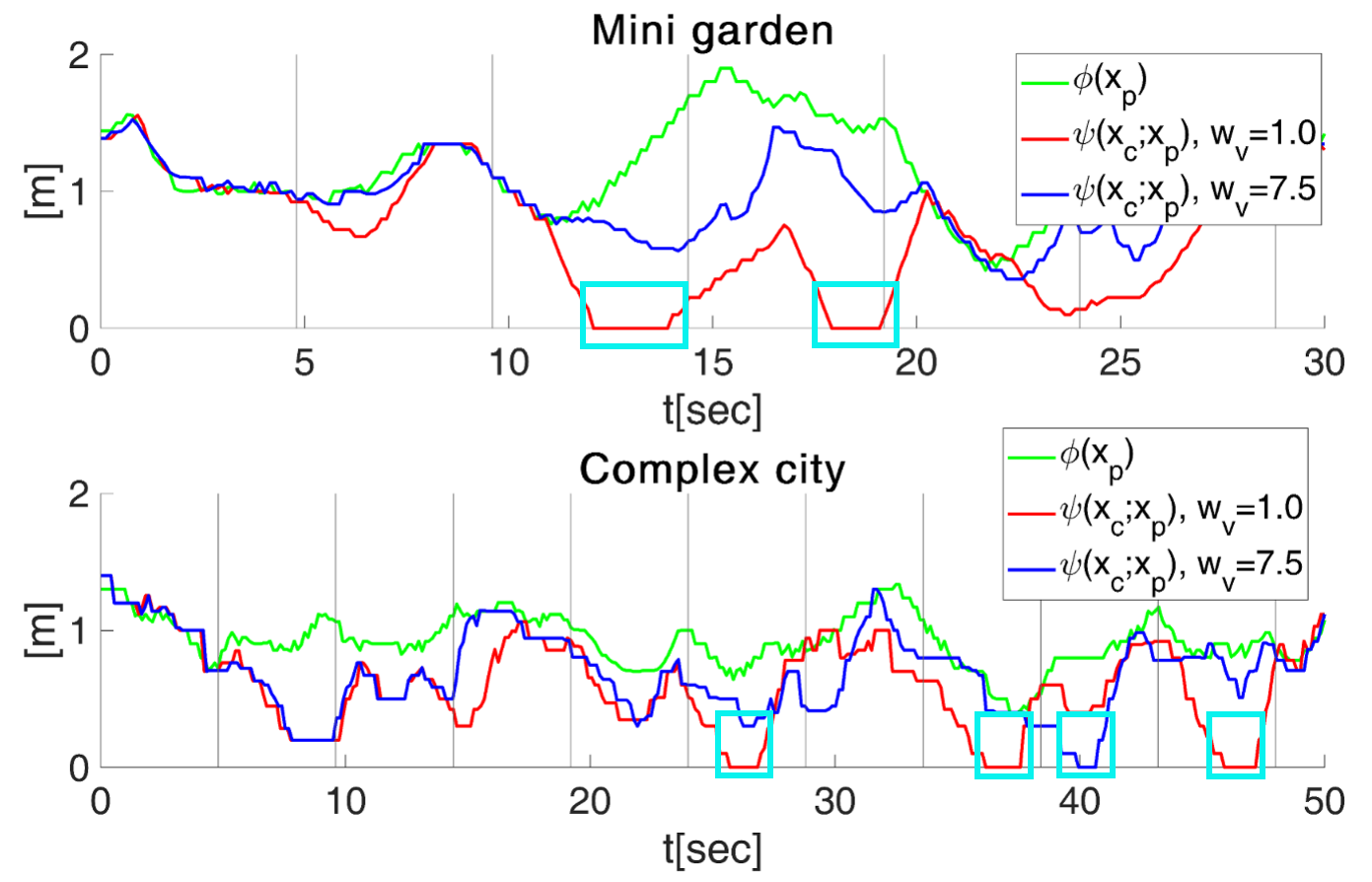}
\caption{The validation result on the visibility performance in the two different environments. The y-axis of the figure denotes $\phi(x_p)$ for target (green) and $\psi(x_c;x_p)$ for the chaser(blue and red for different $w_v$). The cyan dashed boxes represent the duration of occlusion. The small value of $\phi(x_p)$ accounts for the difficulty of maintaining sight for $x_p$.}
\label{fig:data}
\end{figure}

\begin{figure}[h]
\centering
    \includegraphics[width=.5\textwidth]{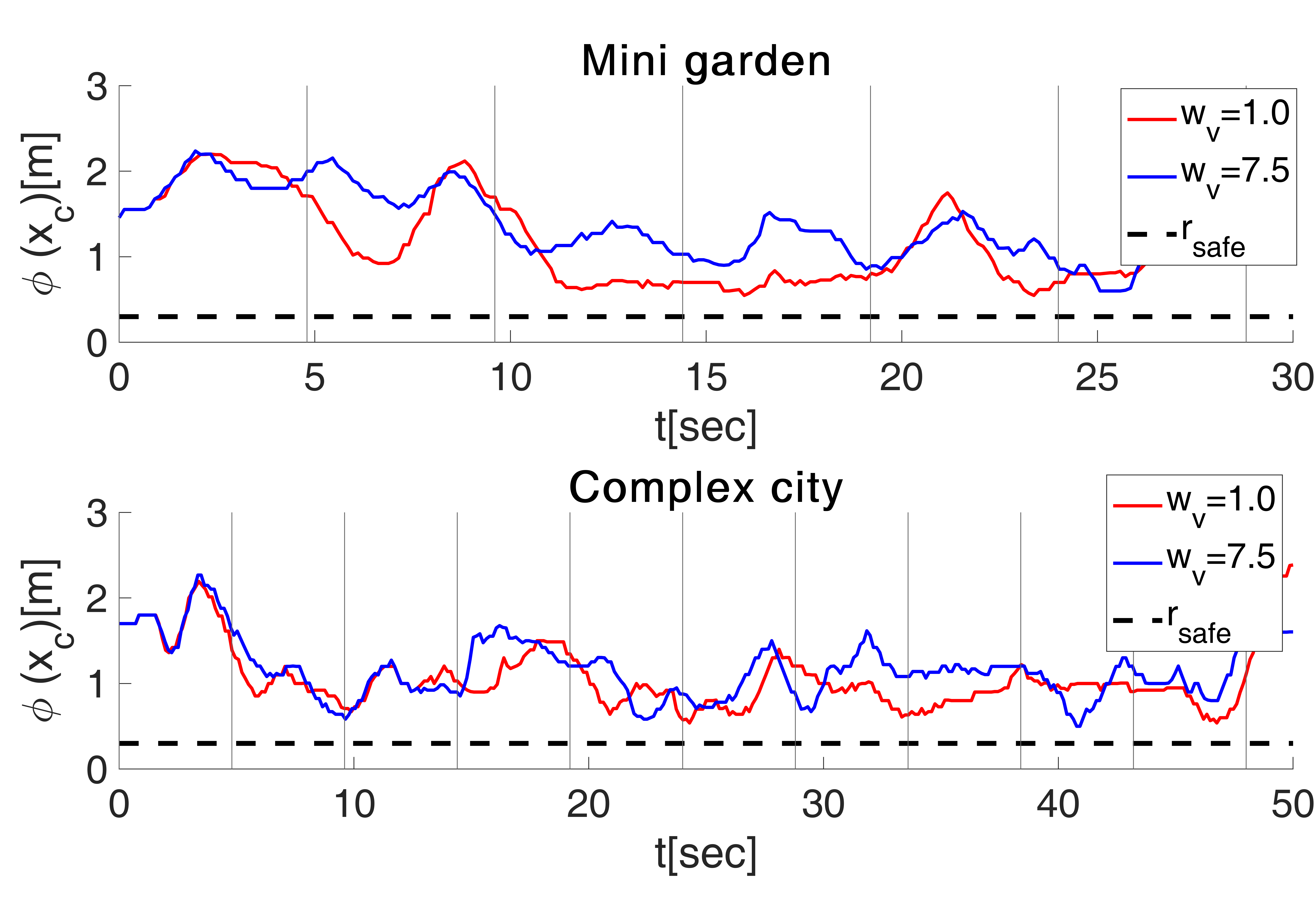}
\caption{The validation result regarding the flight safety of the chaser. For the entire horizon of all the simulations, the safety tolerance $r_{safe} = 0.3$ was satisfied. The average of $\phi(x_c)$ was reduced with smaller $w_v$ as an effort for the shorter travel distance.}
\label{fig:data}
\end{figure}

\setlength\doublerulesep{0.4pt} 
\begin{table}[h]
\begin{center}

\begin{tabular}{c|c c}
\toprule[1pt]\midrule[0.3pt]
\multicolumn{3}{c}{Parameters} \\
\hline
\midrule[0.3pt]
Type & Name & Value\\ \hline
\multirow{2}{*}{Resoluion} & Octomap[m] & res = 0.4 \\
 & $\Omega(x_p)$[m]  & res = 0.8\\
 \hline

\multirow{6}{*}{Optimization } & tracking distance weight & $w_d = 3.4 $ \\
 & waypoint weight& $\lambda = 2$ \\ 
& maximum connection[m] & $d_{max} = 2.0 $\\ 
& polynomial order & $K = 6 $\\ 
& horizon[s] & $H = 5$\\
& \# of corridors & $M = 2$\\
 \hline

\multirow{4}{*}{Tracking spec.}

& desired tracking dist.[m] & $d_{des} = 2.5 $\\ 
& bound of tracking dist. [m] & $1.0\leq \lVert x_c - x_p \rVert \leq 4.0 $\\ 
& safe tol.[m] & $r_{safe} = 0.3 $\\
& tracking elev. $\theta_r$ & $ 20^{\circ}\leq \theta_r \leq 70^{\circ} $\\
\bottomrule[0.3pt]
\bottomrule[0.3pt]
\end{tabular} \caption{Common parameters for simulations}
\label{Table:params}
\end{center}
\end{table}

 In the simulation results in Table \ref{Table:params}, "\# of corridors" means the number of imposed corridor constraints between two knots for a replanning session. The function $\phi(x_p)$ is used to encode the chasing difficulty as it measures density of obstacles along the target path during mission. As can be seen in the table, the total travel distance and the average velocity increased for the higher value of visibility weight $w_v$ while the performance of visibility (i.e. the average of $\psi(x_c;x_p)$) was increased and duration of occlusion was reduced by fraction of five. 
 
 The results for the average computation is also included in Table \ref{Table:result}. To cope with more dense environment as in \textit{complex city}, we imposed the increased number of polynomial segments which leads to the increased computing time for QP. For the graph solving process, in contrary, the computing time was reduced for more dense situation because of the reduced number of feasible edges and nodes in the graph construction. For the both cases, the total planning pipeline runs approximately 10Hz to yield replanning path for the horizon $H=5$ [sec] which is enough for online operation. The safety constraint was satisfied during the entire path with the designed safety clearance as shown in Fig 7.
 
\begin{table}[h]
\centering
\begin{tabular}{cccccc}
\toprule\toprule[1pt]\midrule[0.4pt]
\multicolumn{2}{c}{Data}                                                                                                          & \multicolumn{2}{c}{Mini garden}  & \multicolumn{2}{c}{complex city} \\ \hline
\midrule[0.3pt]
\multicolumn{1}{c|}{\multirow{1}{*}{Planning param.}}                                             & \multicolumn{1}{c|}{$N$}             & \multicolumn{2}{c}{3}            & \multicolumn{2}{c}{4}          \\
\toprule

\multicolumn{1}{c|}{\multirow{2}{*}{Target}}                                               & \multicolumn{1}{c|}{avg. $\phi(x_p)$ {[}m{]}}    & \multicolumn{2}{c}{1.22}         & \multicolumn{2}{c}{0.92}       \\
\multicolumn{1}{c|}{}                                                                      & \multicolumn{1}{c|}{avg. $\rVert \dot{x}_p \rVert$ {[}m/s{]}}  & \multicolumn{2}{c}{0.61}         & \multicolumn{2}{c}{0.58}       \\ \hline
\toprule

\multicolumn{1}{c|}{\multirow{5}{*}{Chaser}}                                               & \multicolumn{1}{c|}{$w_v$}             & 1.0  & \multicolumn{1}{c|}{7.5}  & 1.0            & 7.5           \\ \cline{2-6} 
\multicolumn{1}{c|}{}                                                                      & \multicolumn{1}{c|}{travel dist{[}m{]}} & 22.5 & \multicolumn{1}{c|}{25.5} & 36.53           & 40.1          \\
\multicolumn{1}{c|}{}                                                                      & \multicolumn{1}{c|}{avg. $\lVert \dot{x}_c \rVert${[}m/s{]}}  & 0.75 & \multicolumn{1}{c|}{0.82} & 0.73            & 0.80          \\
\multicolumn{1}{c|}{}                                                                      & \multicolumn{1}{c|}{avg. $\psi(x_c;x_p)${[}m{]}}    & 0.68 & \multicolumn{1}{c|}{0.99} & 0.59           & 0.7           \\
\multicolumn{1}{c|}{}                                                                      & \multicolumn{1}{c|}{occ. duration{[}s{]}}    & 2.5  & \multicolumn{1}{c|}{0}    & 3.0            & 0.6          \\ \hline
\toprule

\multicolumn{1}{c|}{\multirow{5}{*}{\begin{tabular}[c]{@{}c@{}}Comp.\\ time\end{tabular}}} & \multicolumn{1}{c|}{$\phi(x)${[}s{]}}    & \multicolumn{2}{c}{0.057}        & \multicolumn{2}{c}{0.071}      \\
\multicolumn{1}{c|}{}                                                                      & \multicolumn{1}{c|}{$\psi(x;x_p)${[}s{]}}     & \multicolumn{2}{c}{0.001}        & \multicolumn{2}{c}{0.0009}     \\
\multicolumn{1}{c|}{}                                                                      & \multicolumn{1}{c|}{Dijkstra{[}s{]}}     & \multicolumn{2}{c}{0.055}        & \multicolumn{2}{c}{0.0064}     \\
\multicolumn{1}{c|}{}                                                                      & \multicolumn{1}{c|}{QP{[}s{]}}     & \multicolumn{2}{c}{0.0017}       & \multicolumn{2}{c}{0.0037}     \\ \cline{2-6} 
\multicolumn{1}{c|}{}                                                                      & \multicolumn{1}{c|}{total}         & \multicolumn{2}{c}{0.11}         & \multicolumn{2}{c}{0.08}       \\ \hline

\end{tabular}
\caption{Simulation results}
\label{Table:result}

\end{table}

\begin{figure}[h]
 \centering
  \includegraphics[width=0.5\textwidth]{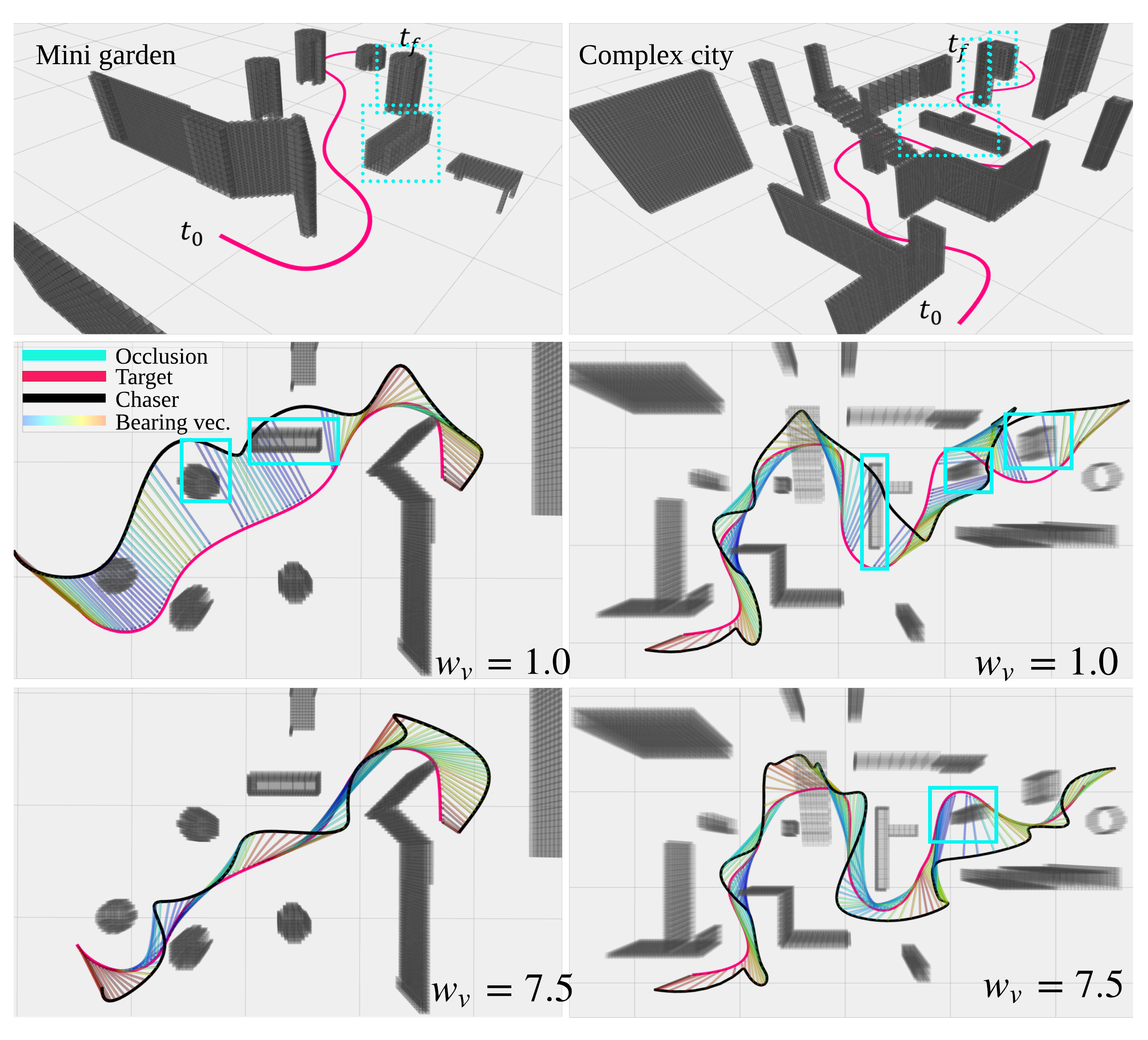}
  
  \caption{ The simulation results: the first column of the figure is the result of a \textit{mini garden }environment and the second one is a   \textit{complex city} with two different values of the weight for visibility $w_v$. The cyan boxes on figures denote the obstacles which caused the occlusion during missions. The relative vector $x_p - x_c$ is plotted with color representing the scale of $\phi(x_c,x_p)$ (red corresponds to the high visibility and blue to low visibility). }
  \label{fig:total result}
\end{figure}

\label{sim}

\newcommand{\LineComment}[1]{\Statex \hfill\textit{#1}}




\section{CONCLUSION}
In this paper, we proposed the cascaded planner for chaser operating in dense environments. By means of the level set-based  metric for visibility of target, the preplnnaer yields a set of  chasing corridors which consider the safety and visibility together with the travel distance. As the following step, dynamically feasible path was generated using convex optimization with numerical stability enjoying the real-time performance. The detailed implementation results were described where the aimed capabilities were satisfied. For the future works, we will improve the proposed pipeline with consideration of the velocity of the moving target. Also, we plan to extend the proposed method to actual hardware platform.     




\end{document}